\documentclass[conference]{IEEEtran}

\ifCLASSINFOpdf
   \usepackage[pdftex]{graphicx}
   \usepackage{subfig}
   \usepackage{multirow}
   \usepackage{amsmath}
   \usepackage{amssymb}
   \usepackage{algorithm,algorithmic}
   \usepackage{url}
   \usepackage{color,soul}
   \usepackage[english]{babel}
   \usepackage{blindtext}
\else
\fi
\begin{document}

\title{WasteNet: Waste Classification at the Edge for Smart Bins}
\author{\IEEEauthorblockN{Gary White,
Christian Cabrera, 
Andrei Palade,
Fan Li,
Siobh\'{a}n Clarke}
\IEEEauthorblockA{School of Computer Science and Statistics,\\ Trinity College Dublin, Dublin, Ireland
\\\{whiteg5, cabrerac, paladea, fali, siobhan.clarke\}@scss.tcd.ie}}
\maketitle

\begin{abstract}
Smart Bins have become popular in smart cities and campuses around the world. These bins have a compaction mechanism that increases the bins' capacity as well as automated real-time collection notifications. In this paper, we propose WasteNet, a waste classification model based on convolutional neural networks that can be deployed on a low power device at the edge of the network, such as a Jetson Nano. The problem of segregating waste is a big challenge for many countries around the world. Automated waste classification at the edge allows for fast intelligent decisions in smart bins without needing access to the cloud. Waste is classified into six categories: paper, cardboard, glass, metal, plastic and other. Our model achieves a 97\% prediction accuracy on the test dataset. This level of classification accuracy will help to alleviate some common smart bin problems, such as recycling contamination, where different types of waste become mixed with recycling waste causing the bin to be contaminated. It also makes the bins more user friendly as citizens do not have to worry about disposing their rubbish in the correct bin as the smart bin will be able to make the decision for them. 
\end{abstract}
\IEEEpeerreviewmaketitle

\section{Introduction}\label{sect:introduction}
Increases in urbanisation and economic development around the world has led to increased waste generation \cite{bandara2007relation, vethaak2016plastic}. In the past century waste production has risen tenfold and by 2025 it is expected to double again \cite{hoornweg2012waste}. This waste is often not recycled properly and is usually either dumped in a landfill or incinerated \cite{king2006reducing}. Landfills, such as Laogang in Shanghai, Sudokwon in Seoul, Jardim Gramacho in Rio de Janeiro and Bordo Poniente in Mexico City are among the world's largest and show that this is a global problem \cite{robinson2007composition}. Waste incinerators are being used to deal with the waste problem, but there are a number of concerns over fine particle emission, which can lead to a number of associated health risks as well as contributing to global warming \cite{zhou2015enrichment, naroznova2016global}. This can lead to the accumulation of polluting materials that threaten the environment \cite{da2016nano, hopewell2009plastics, cozar2015plastic} as well as human \cite{kim2015review, wright2017plastic} and animal health \cite{mattsson2018nanoplastics, derraik2002pollution}. Waste products can be classified based on various factors, such as consumption, production, chemical and physical properties to allow for effective reuse and recycling. The ability to reduce and recycle waste material more effectively not only reduces the impact on the environment, but is also more cost effective \cite{zaman2016comprehensive, xu2018global}. 

\begin{figure}[!tb]
	\centering
	\includegraphics[width=\linewidth]{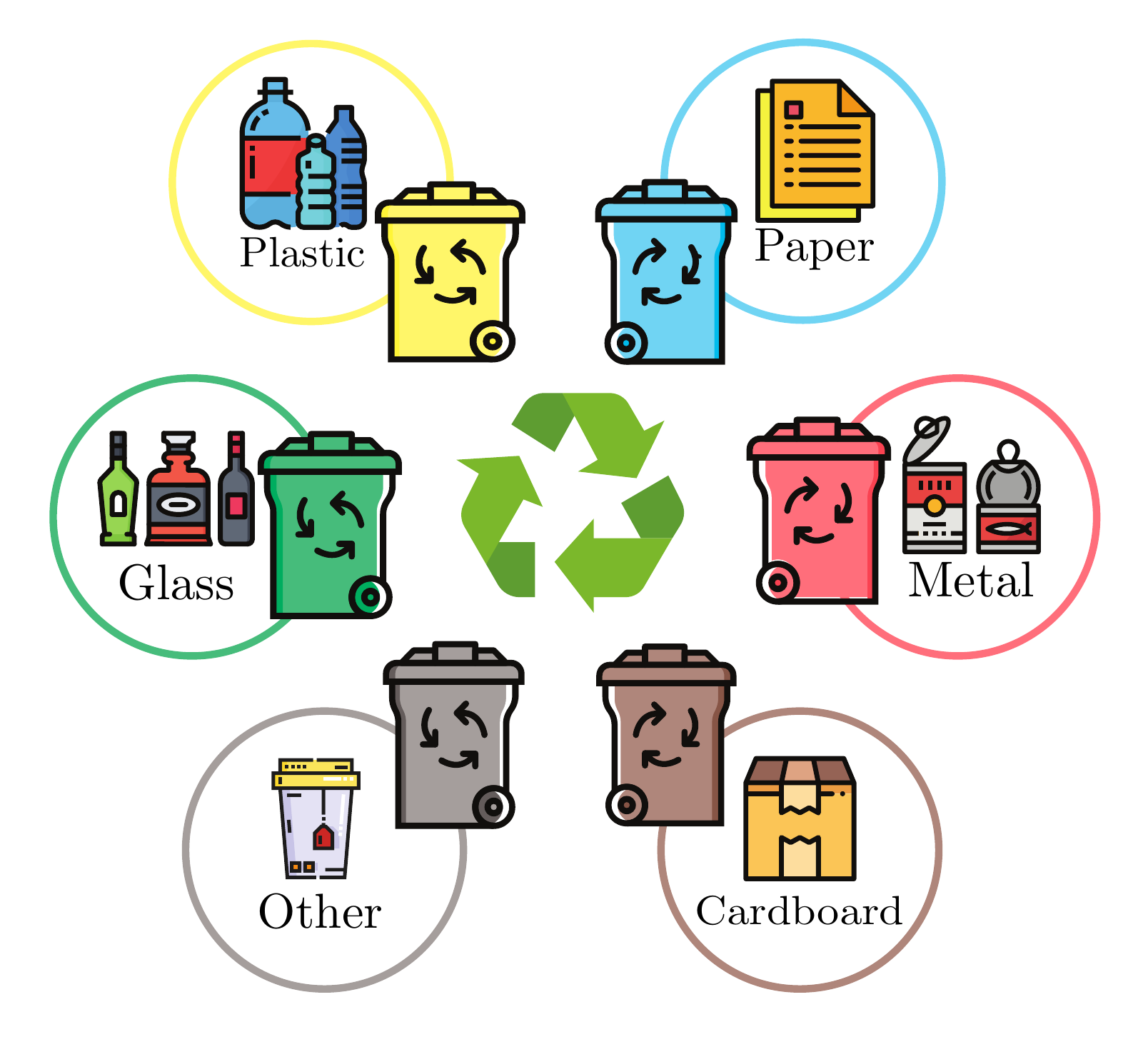}
	\caption{Waste Categories}
	\label{fig:waste_categories}
\end{figure}

Waste classification can happen at any stage in the waste management pipeline, but is it better to happen when the waste is initially being disposed of to avoid potential recyclable materials becoming contaminated. Solar powered compaction bins have become increasingly popular in smart cities and campuses as they can allow for eight times the capacity of a traditional bin due to the compaction and can also send an alert when the bin is full \cite{poss2006solar}. However, these bins do not have any waste classification mechanism, which can lead to a recycling bin becoming contaminated \cite{andrews2013comparison}. General waste can be classified into the categories shown in Figure \ref{fig:waste_categories} to allow for more efficient recycling. The most common contaminants in an urban environment are napkins, plastic food wrapping, plastic bags, coffee cups, coffee sleeves, rubber gloves and plastic medical waste \cite{heathcote2010conducting, sajjad2020analyses}. A number of schemes have been tried to inform and better educate the public on the categories and benefits of waste classification in different demographics \cite{mourad2016recycling}. An automated classification mechanism would solve this problem and encourage user engagement as they would not be worried about placing an item in a recycling bin that could possible contaminate it. 

Deep neural networks have shown increased accuracy in a number of recent benchmark competitions in machine learning and pattern recognition \cite{8949510}. We can frame the waste classification problem as an image classification problem where we  have a camera that takes an image of the waste and uses a deep neural network to identify what class the waste belongs to. For image-based tasks standard deep neural networks are hard to train and encounter problems when trying to scale \cite{ciresan2011flexible}. Traditional dense connections do not have the translation-invariance property, so any slight change in the size or placement of an image would be difficult to detect. Convolutional neural networks (CNNs) solve this problem by using multiple hidden convolutional layers that extract high level features. Convolutional layers consist of learnable parameters, called filters that go through the whole image i.e., it steps across the width, length and colour channels of the image and calculates the inner product of the input and the filter. This leads to a feature map of the filter, which is used to classify the images. 

Another building block of convolution networks, is a pooling layer, which operates on the feature maps. The pooling layer reduces the spatial size of the representation to reduce the chance of overfitting and cuts down the number of parameters. Max pooling is an approach that divides the space into individual regions and picks the maximum value for each region. CNNs also make use of the ReLu activation function of the form $f(x) = max(0, x)$ as this leads to faster training without effecting the accuracy of the network \cite{krizhevsky2012imagenet}. The combination of these recent advancements has allowed CNNs to be used in a range of computer vision tasks, such as: face detection \cite{jiang2017face}, license plate recognition \cite{polishetty2016next}, multi-label classification \cite{wang2016cnn}, noise recognition \cite{cao2018urban}  and object detection \cite{chen2017multi}. 

In this paper we present WasteNet, a deep neural network model that improves waste classification accuracy. This model is tested on a Jetson Nano edge device that can easily be deployed at the edge to allow waste classification in smart bins. Our model achieves an improved state of the art accuracy score of 97\% on the TrashNet dataset \cite{yang2016classification}. This is a large improvement on the original SVM approach that achieved an accuracy of 63\% and CNN approach that achieved an accuracy of 22\% \cite{yang2016classification}. A recent survey of automated image-based waste classification papers, has shown a state of the art accuracy on the TrashNet dataset is 88.42\% \cite{10.1007/978-3-030-19651-6_41}.

The paper is organised as follows, Section \ref{sect:related_work} presents the related work that has been conducted on waste classification. Section \ref{sect:design} presents the design of the deep learning model and the categories of waste that it uses when making the classification decision. Section \ref{sect:experimental_setup} presents the experimental setup that was used to evaluate our proposed deep learning model against other state of the art approaches and Section \ref{sect:results} presents the results of those experiments. Section \ref{sect:conclusion} concludes the paper and presents some future work.

\section{Related Work}\label{sect:related_work}
Traditional approaches to waste classification have focused on the physical properties, such as the weight, form (solid, liquid, aqueous or gaseous) and the type of process that has generated the waste \cite{zekkos2010physical}. These properties can then be used by a machine learning \cite{kuritcyn2015increasing} or fuzzy set theory \cite{musee2008new} algorithm to classify the waste according to the set categories. For incinerators, the classification of waste can be for the thermal conversion in waste-to-energy research \cite{zhou2015classification}. In vitro testing, such as Microtex, which uses bioluminescent bacteria to detect hazardous waste has also been conducted \cite{bitton1994evaluation}. Alternative techniques have also been used to estimate waste generation using GIS-based modelling \cite{karadimas2008gis} and multi-agent systems to simulate household behaviour \cite{meng2018multi}.  

In recent years, visual methods using either images in the visible \cite{sakr2016comparing} or infrared \cite{kuritcyn2015increasing} spectrum have become popular. Image classification has become a major research area thanks to the release of large publicly available datasets, such as ImageNet \cite{imagenet_cvpr09}, which has led to the development of large deep neural network models. Recent computer vision approaches have been used to localise and classify waste on the streets \cite{rad2017computer, 8704340}. This allows street cleaning equipment to find the areas that have the most waste and to focus on cleaning those areas. Computer vision approaches have also been applied to internal cleaning robots to identify and detect the type of waste that the robot is about to clean on the floor \cite{ramalingam, torrey2010transfer, ramalingam2018cascaded}. This can also be applied at a larger urban scale with Urban Swarms for autonomous waste management \cite{alfeo2019urban}. At the moment these urban swarms are proposed to allow for the efficient collection of waste throughout the city from bins, but could easily be extended to contain waste localisation and classification technology to allow the agents in the swarm to automatically pick up and classify waste as they move through the streets. 

Previous image-based waste classification systems have focused on specific subsections of waste classification, such as classifying plastic into: polyethylene terephthalate, high-density polyethylene, polypropylene and polystyrene \cite{10.1007/978-3-030-20518-8_30} or focusing on household waste \cite{8802522}. Our approach is designed to be deployed at the edge and so would serve as a first pass classification mechanism that first divides the waste into the classes defined in the TrashNet dataset: plastic, glass, paper, metal, cardboard and other. This first pass then allows for further classification of specific classes, such as plastic at dedicated plastic recycling centres. Other waste classification approaches make use of other devices, such as inductors and bridge sensors to influence the classification of waste \cite{chu2018multilayer}. In this paper we focus on the use of image classification for the classification of the waste at the edge of the network.

\begin{figure}
\centering
\includegraphics[width=\linewidth]{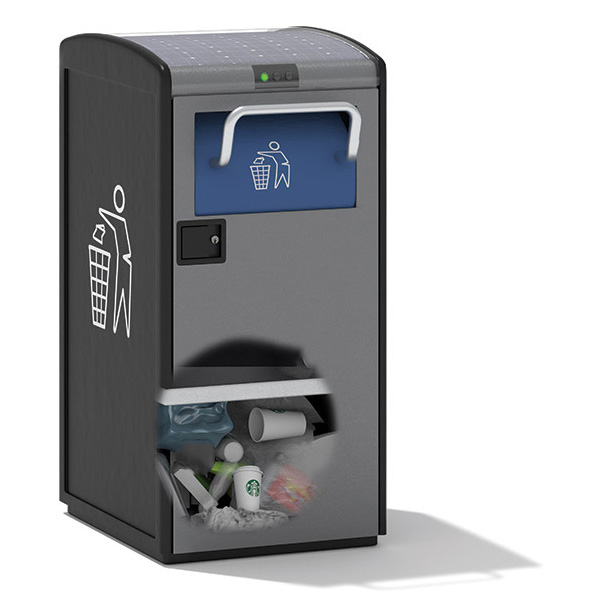}
\caption{Urban Smart Bin}
\label{fig:bigbelly}
\end{figure}

\section{WasteNet Design}\label{sect:design}
In urban environments self-compacting bins have become popular because due to a compression mechanism the can hold eight time more waste than a typical bin \cite{poss2006solar}. However, these bins compact all the waste together without classification, which can lead to recyclable material becoming contaminated. In this paper we propose using a deep neural network-based model to classify the waste as it is added to the bin. Figure \ref{fig:bigbelly} shows how these bins accept waste in a separate container at the top, which could easily be fitted with a camera to allow for images of the waste to be taken. These images would then be fed to our model, which would select the correct container for the waste to be disposed in to ensure that it would be recycled properly. These bins are also fitted with solar panels allowing for a low power edge device, such as a Nvidia Jetson Nano to run the model locally. The transparent section at the bottom of the bin shows how plastic, cardboard and paper are currently mixed and squashed together at the bottom of the bin, without waste classification. The remainder of this section describes the design of the WasteNet neural network model. 

\subsection{Transfer Learning}
The design of the WasteNet model uses transfer learning to leverage knowledge from a source task \cite{torrey2010transfer, 9007504}. We leverage knowledge from models trained for a general image classification task on the ImageNet dataset \cite{imagenet_cvpr09}. Transfer learning can also provide a number of benefits, such as improving baseline performance, speeding up overall model development and training time and also getting overall improved model performance compared to building the model from scratch. This is especially important in deep learning where models can have a very long and intensive training time. 

\begin{figure}
	\centering
	\includegraphics[width=\linewidth]{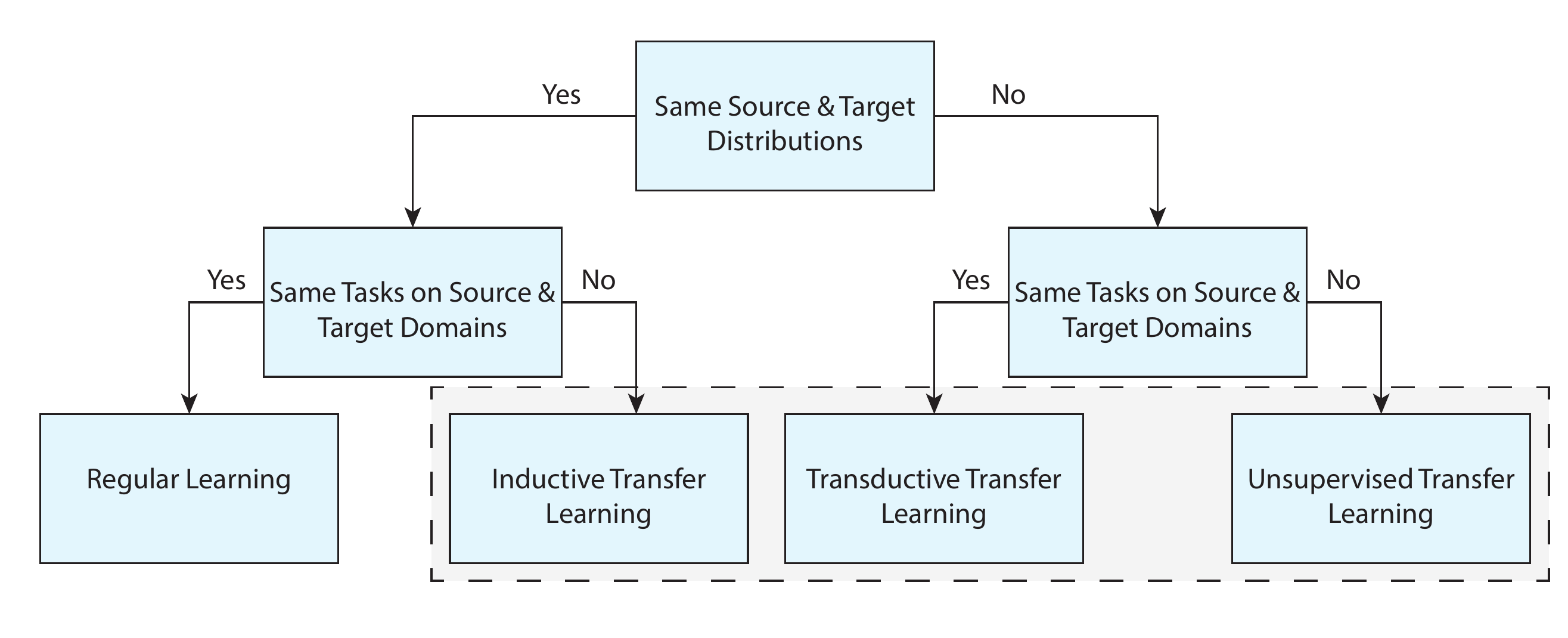}
	\caption{Transfer Learning Categories}
	\label{fig:transferlearning}
\end{figure}

Figure \ref{fig:transferlearning} shows the different categories of transfer learning based on the relationship between the source and target distributions. The most simple case is regular learning where the source and target have the same distributions and are required to perform the same tasks. When the source and target have the same distribution or are in the same domain but the tasks that they are required to perform are different this is called inductive transfer learning. This category can further be broken down depending upon whether the source domains contain labelled data or not: if a lot of labelled data in the source domain are available then it is multi-task learning and if there are no labelled data from the source domain then it is self-taught learning \cite{han2011data}. When the source and target distribution are not the same but the tasks are similar it is called transductive transfer learning. In this situation, no labelled data in the target domain are available, while a lot of data in the source domains are available. The final category is unsupervised transfer learning, where there is a difference in both the source and target distribution and tasks. This category focuses on solving unsupervised tasks in the target domain such as clustering \cite{dai2008self} and dimensionality reduction \cite{wang2008transferred}, with no labelled data available in the source and target domains in training. A linear cost function can be used to minimise the difference between the source and target domain distribution in unsupervised transfer learning \cite{Nejatian2019, Pirbonyeh2019}. In this paper we focus on inductive transfer learning, where the original model is trained on the same source and target distribution, but performs a different tasks of waste classification. 

\subsubsection{Feature Extraction}
\begin{figure}[!htb]
	\centering
	\subfloat[][VGG-16 Model]{\includegraphics[width=.16\textwidth]{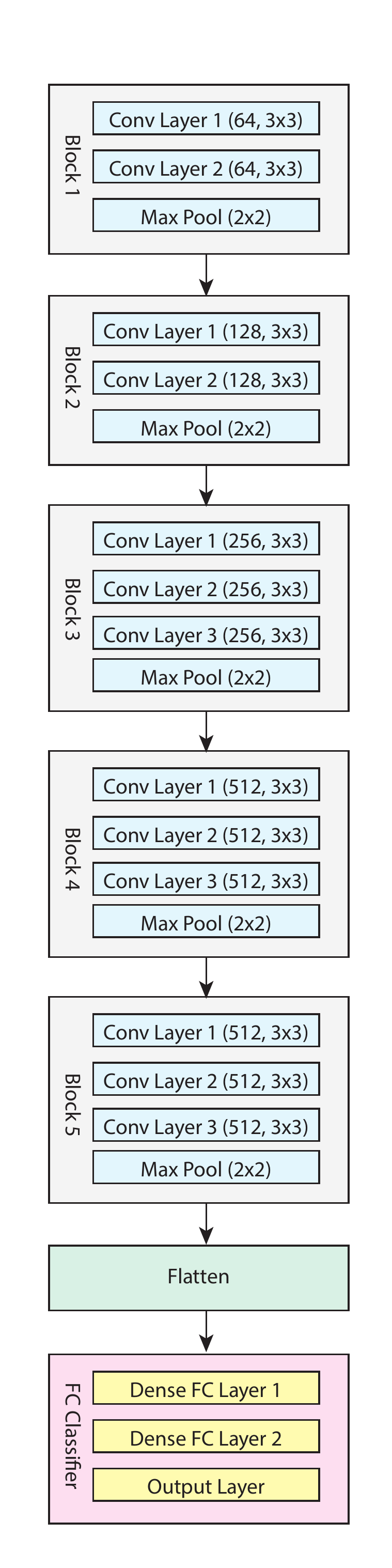}\label{fig:vgg}}\hfill
	\subfloat[][\centering VGG-16 Model \newline Feature Extractor]{\includegraphics[width=.16\textwidth]{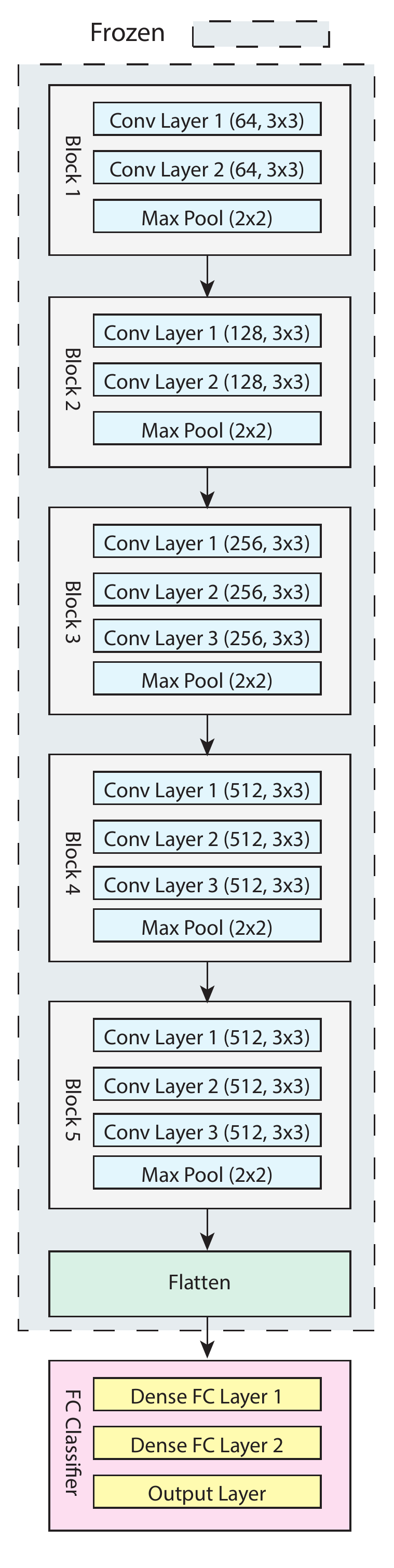}\label{fig:vgg_feature}}\hfill
	\subfloat[][\centering VGG-16 Model \newline Fine Tuning]{\includegraphics[width=.16\textwidth]{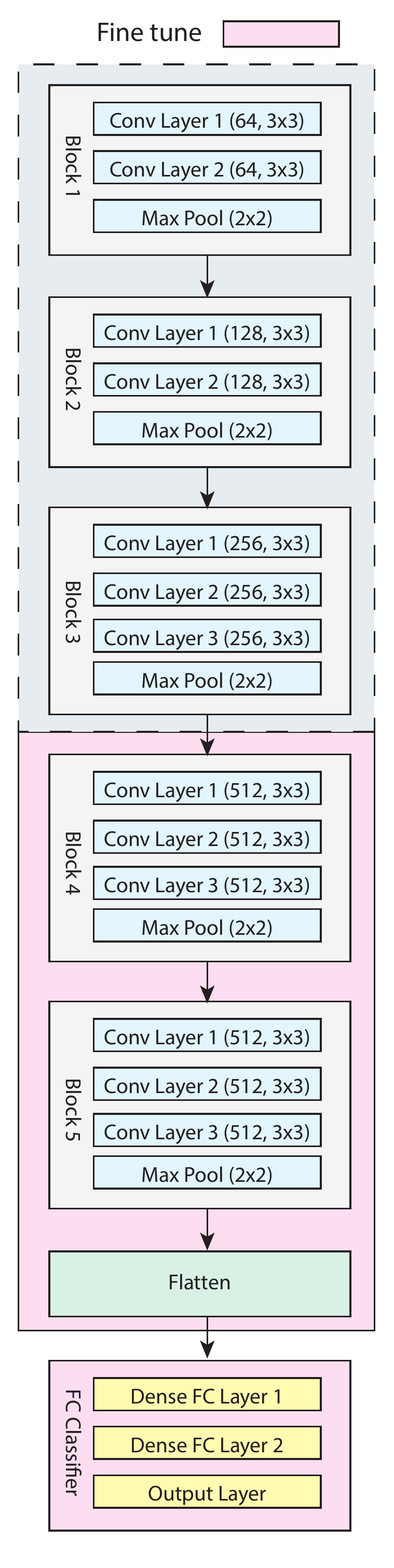}\label{fig:vgg_fine}}
	\caption{Deep Transfer Learning Approaches}
	\label{fig:vgg_transfer}
\end{figure}

Deep learning models are layered architectures that learn different features at different layers. These layers are finally connected to a final layer, which is usually fully connected in the case of classification to get the final output class. This layered architecture allows us to utilize pre-trained networks without a final layer as a fixed feature-extractor for other tasks. Figure \ref{fig:vgg_transfer} shows the different transfer learning approaches that can be applied to this task. Figure \ref{fig:vgg} shows the VGG-16 model that can be downloaded directly to an edge device \cite{2014arXiv1409.1556S}. Figure \ref{fig:vgg_feature} shows the VGG model as a feature extractor where we freeze (fix weights and don't train) all the blocks of convolutions layers and the flattening layer in the dashed blue box. We only update the fully connected classifier block at the end of the model. This allows the new model in this case to transform the image from a new domain task into a large dimension vector based on hidden states, thus enabling us to extract features from a new domain task using the source domain. This is one of the most widely used methods of transfer learning for deep neural networks. 

\subsubsection{Fine Tuning}
In fine tuning, the weights of the last few layers of the network are updated as shown in Figure \ref{fig:vgg_fine} with a pink box and trained as well as the fully connected layers at the end of the model, for the classification task. This means that this method is slightly more resource intensive as we have to train some of the previous layers. As deep neural networks are layered with the initial layers capturing the most basic features, such as edges and the later layers capturing more specific details about the task, we can freeze some of the first blocks and update the later ones. In Figure \ref{fig:vgg_fine} we freeze the first three layers in the dashed blue box, while fine tuning the last two blocks to suit the task. This allows us to use the knowledge in terms of the overall architecture of the network and use its states as the starting point for our retraining step allowing us to achieve better performance in less time. 

One of the problems with updating a model using fine-tuning is that some of the parameters in the non-frozen layers have to be updated to solve the new problem. This can overwrite parameters that the network has learned before leading to ``catastrophic forgetting" of the knowledge that was previously acquired \cite{french1999catastrophic, 2017arXiv170802072K}. Previous approaches to preventing catastrophic forgetting have used an ensemble of neural networks. When there is a new task, the algorithm creates a new network and shares the representation between the tasks \cite{Lee:2016:DDL:3060832.3060854, rusu2016progressive}. However, these approaches are not suitable for the edge due to the space and complexity restrictions as the number of networks increases linearly with the number of new tasks to be learned. 

\subsubsection{Hybrid Tuning}
In this paper we propose a hybrid transfer learning approach that combines the benefits of feature extraction and fine-tuning. Our hybrid approach has two stages, in the first pre-training stage the base network is used as a feature extractor by freezing the lower layers of the network and only updating the weights of the top layer. Once the loss function begins to stabilise and the network has reached a high level of accuracy with the new top layer of the network, the remaining layers of the network are gradually unfrozen. The remaining layers of the network are trained using discriminative layer training, where a different learning rate is applied to each layer. This allows us to apply a lower learning rate to the low-level layer representations and adapt the weights of the higher-level layers faster as they contain more domain specific information. 

This gradual unfreezing rather than fine-tuning all the layers at once reduces the risk of catastrophic forgetting. We first unfreeze the top layer as this contains the least general knowledge and update the weights \cite{yosinski2014transferable}. We then unfreeze the next lower frozen layer and repeat using the updated learning rate, until we fine-tune all layers with convergence at the last iteration. To generate the WasteNet model we apply our hybrid tuning approach to a DenseNet model that has been trained on ImageNet \cite{krizhevsky2012imagenet}. 

\subsection{Data Augmentation}
Generating artificial data based on existing observations is a technique in machine learning to control overfitting, improve model accuracy and generalisation \cite{bloice2017augmentor}. The idea behind this technique also known as data augmentation is that we follow a set process, taking existing data, such as images from our training set and applying some image transformation operations on them, such as translation, zooming, shearing and rotation to produce a new set of alternative images. The randomness of the process means that we do not get the same images each time. This helps to stop the deep learning model from overfitting on the local training data. In our experiments we use the ImageDataGenerator class\footnote{\label{keras}https://keras.io/preprocessing/image/\#imagedatagenerator-class} from Keras to provide a number of transformations for generating new images, such as: zooming, rotation, translation, randomly flipping images horizontally, and filling new pixels with their nearest surround pixels.

\section{Experimental Setup}\label{sect:experimental_setup}

\subsection{Dataset}
\begin{figure}[!tb]
	\centering
	\subfloat[][Cardboard]{\includegraphics[width=.16\textwidth]{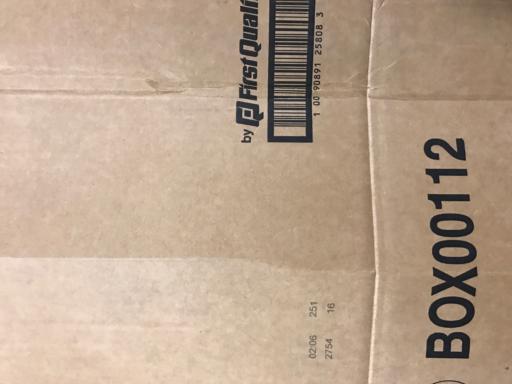}}\hfill
	\subfloat[][Glass]{\includegraphics[width=.16\textwidth]{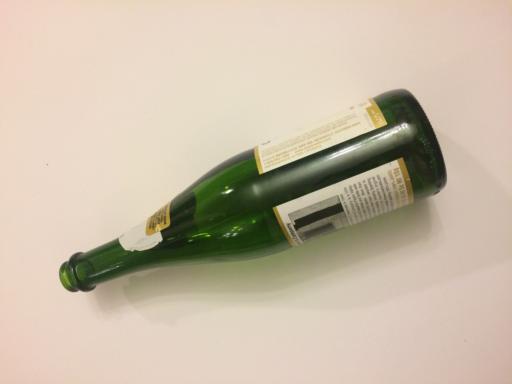}}\hfill
	\subfloat[][Metal]{\includegraphics[width=.16\textwidth]{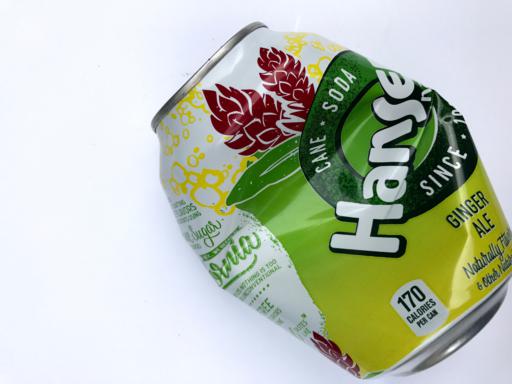}}\par
	\subfloat[][Paper]{\includegraphics[width=.16\textwidth]{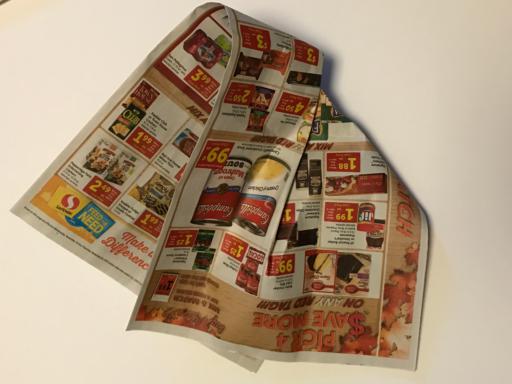}}\hfill
	\subfloat[][Plastic]{\includegraphics[width=.16\textwidth]{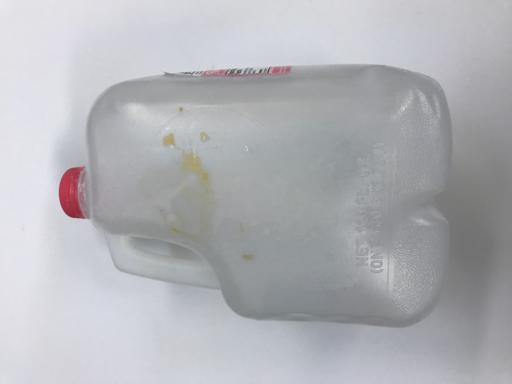}}\hfill
	\subfloat[][Other]{\includegraphics[width=.16\textwidth]{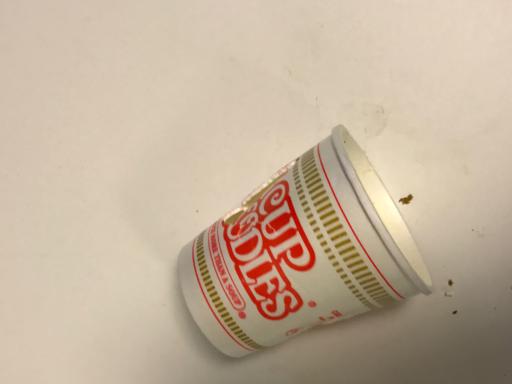}}\par
	\subfloat[][Cardboard]{\includegraphics[width=.16\textwidth]{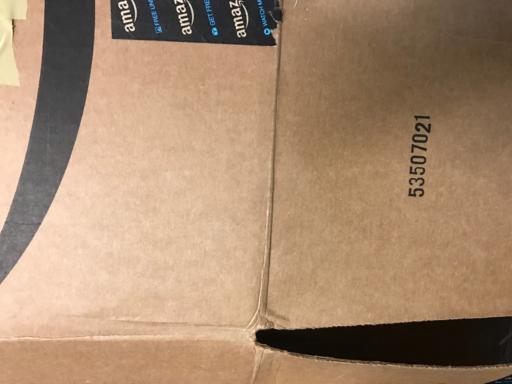}}\hfill
	\subfloat[][Glass]{\includegraphics[width=.16\textwidth]{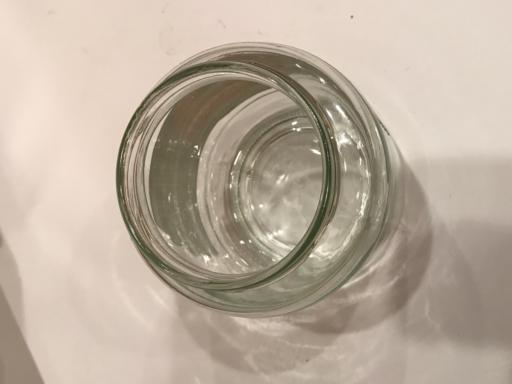}}\hfill
	\subfloat[][Metal]{\includegraphics[width=.16\textwidth]{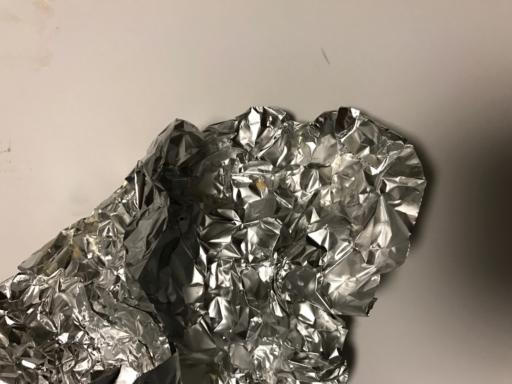}}
	\caption{Images of Waste Categories} 
	\label{fig:waste_categories_images}
\end{figure}

To evaluate our model we use the TrashNet dataset \cite{yang2016classification}. The dataset contains multiple classes: plastic, paper, glass, metal, cardboard and other. It also contains a range of images of waste in different orientations and positions as seen in Figure \ref{fig:waste_categories_images}. The exposure and lighting for each of the photos is also varied in the dataset. Each image is resized to 512x384 with 3 colour channels. The original dataset is 3.5GB in size, containing 2527 images in total. The number of images in each class varies with 594 paper, 501 glass, 410 metal, 482 plastic, 493 cardboard and 137 for other materials. For the evaluation, the images are split into a train, validation and test set with a 50, 25, 25 split. 

\subsection{Metrics}
We use a number of classification metrics to comprehensively show how the classification models perform. The metrics are based on the number of True Positives (TP), False Positives (FP), True Negatives (TN) and False Negatives (FN) and are defined as:

\begin{equation}Accuracy = \frac{True\:Positive + True\:Negative}{TP + TN + FP + FN}\end{equation}
\begin{equation}Precision = \frac{True\:Positive}{True\:Positive + False\:Positive}\end{equation}
\begin{equation}TPR/Recall = \frac{True\:Positive}{True\:Positive + False\:Negative}\end{equation}
\begin{equation}F1 = \frac{2 \cdot Precision\cdot Recall}{Precision+ Recall}\end{equation}

\subsection{Performance Comparison}
We compare the results of our classification approach against a range of other state of the art deep learning models. For each model we also use a specific instance of the models with a different numbers of layers. This can be seen in the ResNet model, where we use ResNet34, Resnet50 and ResNet101 to evaluate the impact that model size has on the final accuracy metrics \cite{DBLP:journals/corr/HeZRS15}. We also evaluate against other state of the art deep neural network models, such as DenseNet \cite{2016arXiv160806993H}, SqeezeNet \cite{DBLP:journals/corr/IandolaMAHDK16}, AlexNet \cite{DBLP:journals/corr/Krizhevsky14} and
VGG \cite{2014arXiv1409.1556S}.

\section{Results}\label{sect:results}

\subsection{Classification Accuracy}
Table \ref{tab:transfer_learning_results} shows the results for each of the trash classification methods. We evaluate our WasteNet approach against 10 other state of the art deep neural networks using 4 different metrics for a comprehensive analysis. We can see that the WasteNet approach shows an increase in accuracy compared to existing state of the art approaches, such as DenseNet169 and ResNet101. There is a 0.017 difference in accuracy from WasteNet compared to the next most accurate model. There is a large difference in accuracy between the WasteNet and AlexNet model, with the AlexNet model achieving an accuracy of 0.784, which is 0.186 less than WasteNet. The precision and recall metrics follow a similar pattern with WasteNet achieving the best results. The F1 Score, which is a combination of the precision and recall values are also highest for the WasteNet model. 

\begin{table}[!tb]
	\centering
	\resizebox{0.5\textwidth}{!}{%
		\begin{tabular}{llllll}
			\hline 
			Model & Accuracy & Precision & Recall & F1 Score \\ \hline
			WasteNet & 0.970 & 0.970 & 0.970 & 0.970 \\
			DenseNet169 & 0.953 & 0.954 & 0.953 & 0.953 \\
			DenseNet121 & 0.942 & 0.942 & 0.942 & 0.941 \\
			ResNet101 & 0.939 & 0.939 & 0.939 & 0.938 \\
			ResNet50 & 0.937 & 0.937 & 0.937 & 0.937 \\
			VGG16 & 0.928 & 0.927 & 0.928 & 0.927 \\
			ResNet34 & 0.918 & 0.918 & 0.918 & 0.918 \\
			DenseNet161 & 0.913 & 0.917 & 0.913 & 0.912 \\
			VGG19 & 0.907 & 0.908 & 0.907 & 0.907 \\
			SqueezeNet & 0.811 & 0.814 & 0.811 & 0.812 \\
			AlexNet & 0.784 & 0.786 & 0.784 & 0.782 
			\\ \hline
		\end{tabular}%
	}
	\caption{Waste Classification Accuracy}
	\label{tab:transfer_learning_results}
\end{table}

\begin{figure}[!htb]
	\centering
	\includegraphics[width=\columnwidth]{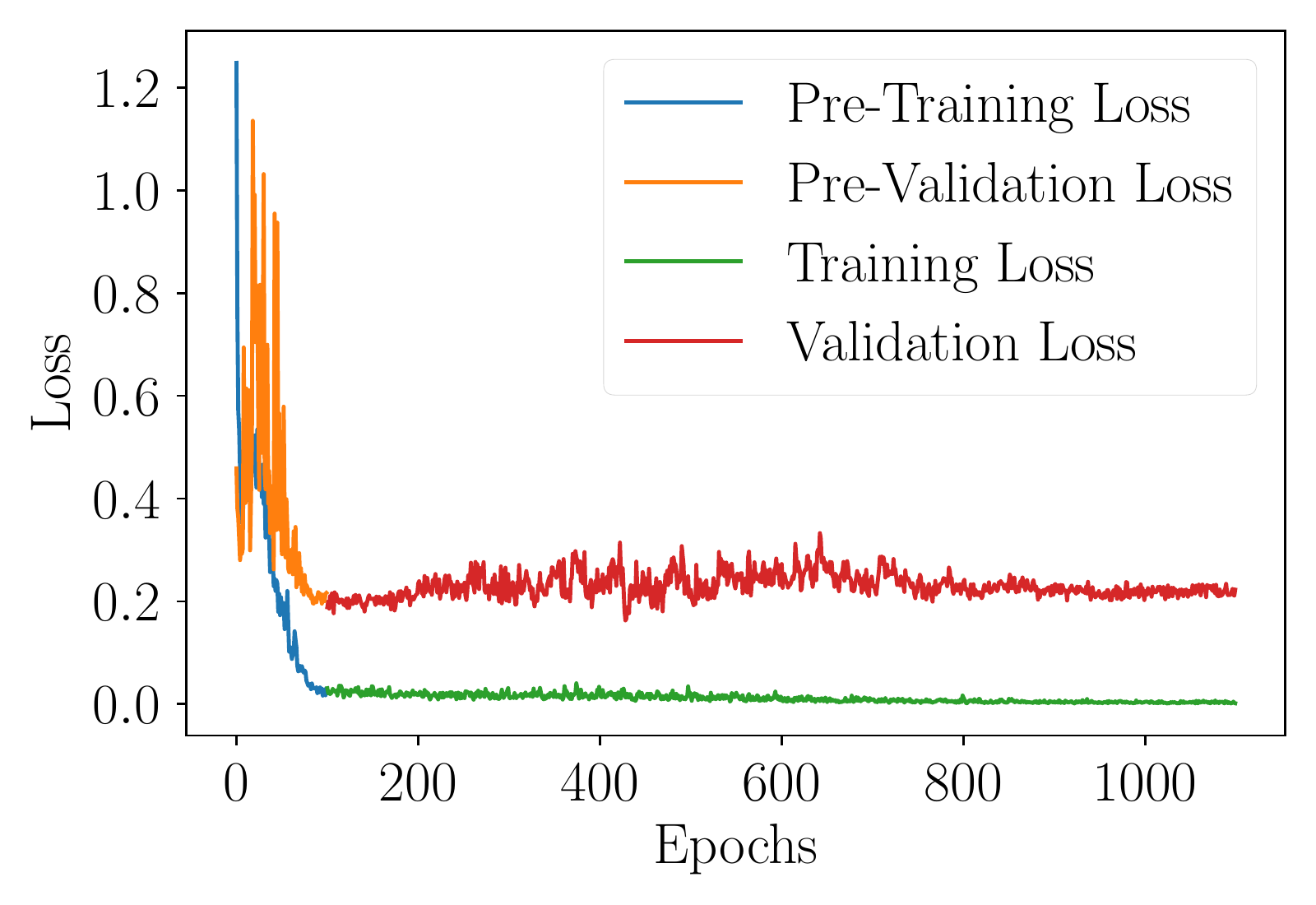}
	\caption{Loss Over Epochs}
	\label{fig:loss}
\end{figure}

Figure \ref{fig:loss} shows the training and validation loss of the network as it is being trained. The orange and blue lines show the pre-training and pre-validation loss. This is when only the top layer of the network is being trained and the remaining layers are frozen for the first 100 epochs. We can see that during this period the loss function is high as the new top layer of the network is learning to classify the waste material. 

At around 100 epochs the loss value begins to stabilise and falls to a lower amount. This is when the second stage of training begins, where we gradually unfreeze the remaining layers of the network and begin training the layers at different learning rates. The training loss at this stage is shown by the green line and the validation loss is shown be the red line. We can see how gradually unfreezing the layers reintroduced some variation especially in the validation loss function. This can be helpful to avoid the network becoming stuck in local maximums or minimums. After 1000 epochs both the validation and training loss remain quite stable. 

\begin{figure}[!tb]
	\centering
	\includegraphics[width=\columnwidth]{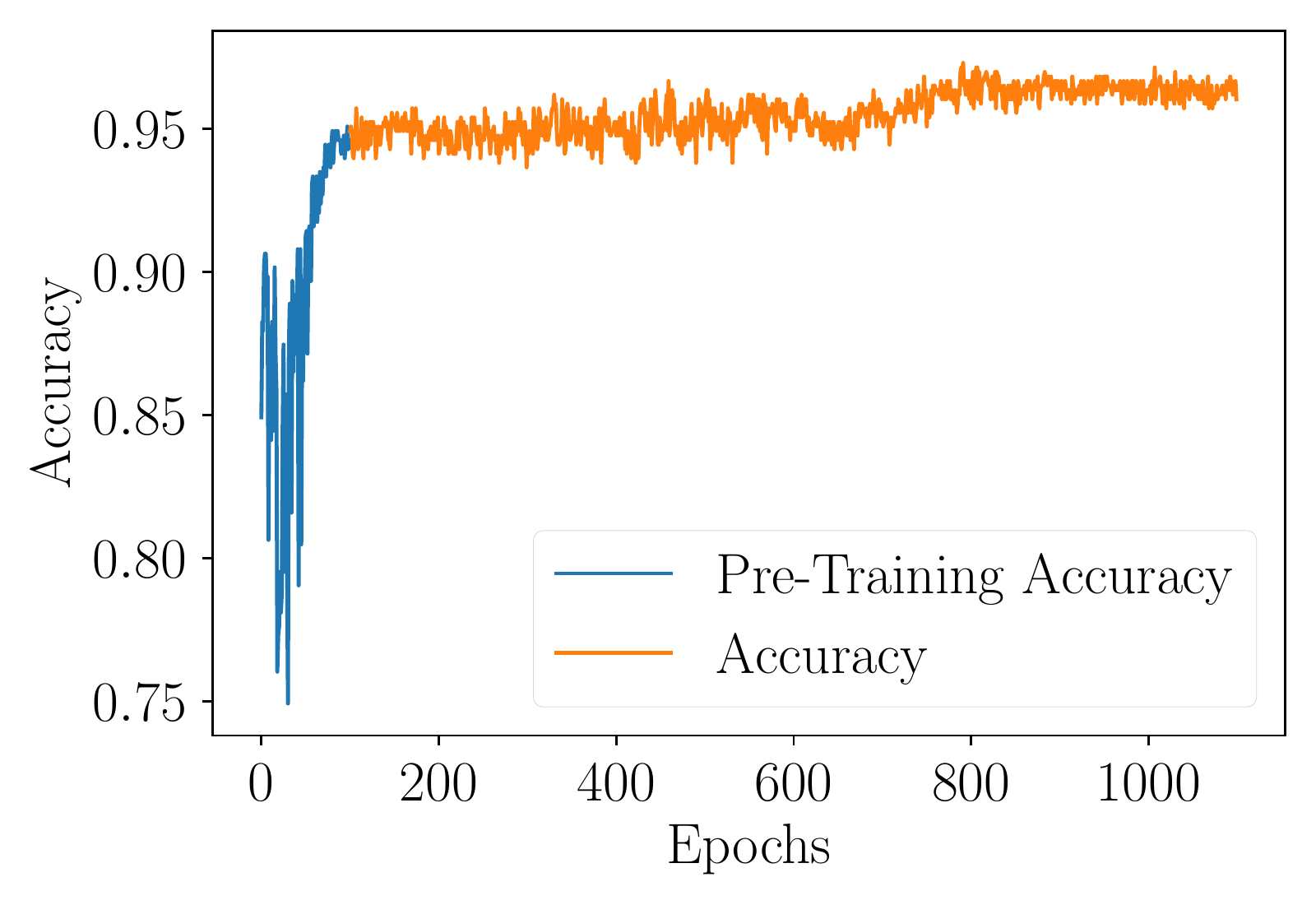}
	\caption{Accuracy Over Epochs}
	\label{fig:accuracy}
\end{figure}

Figure \ref{fig:accuracy} shows the pre-training and training accuracy. The blue line shows the network during pre-training, when the top layer of the network is trained and all the others remain frozen. The orange line shows the network when the remaining layers are being gradually unfrozen and trained at different learning rates. Similar to Figure \ref{fig:loss} there is a lot of variation in the pre-training stage, when the network is beginning to classify the waste. We can see that accuracy in this stage varies a lot starting at around 0.9 and dropping to 0.75. Once the network begins to stabilise the accuracy at around 100 epochs the remaining layers are gradually unfrozen to allow for further accuracy improvement. The accuracy still varies a lot with a noticeable increase around 700 epochs, when there is a lot of variation in the validation loss function in Figure \ref{fig:loss}. After this increase the network accuracy remains quite stable. 

\subsection{Classification Confusion Matrix}
\begin{figure}[!tb]
	\centering
	\includegraphics[width=\columnwidth]{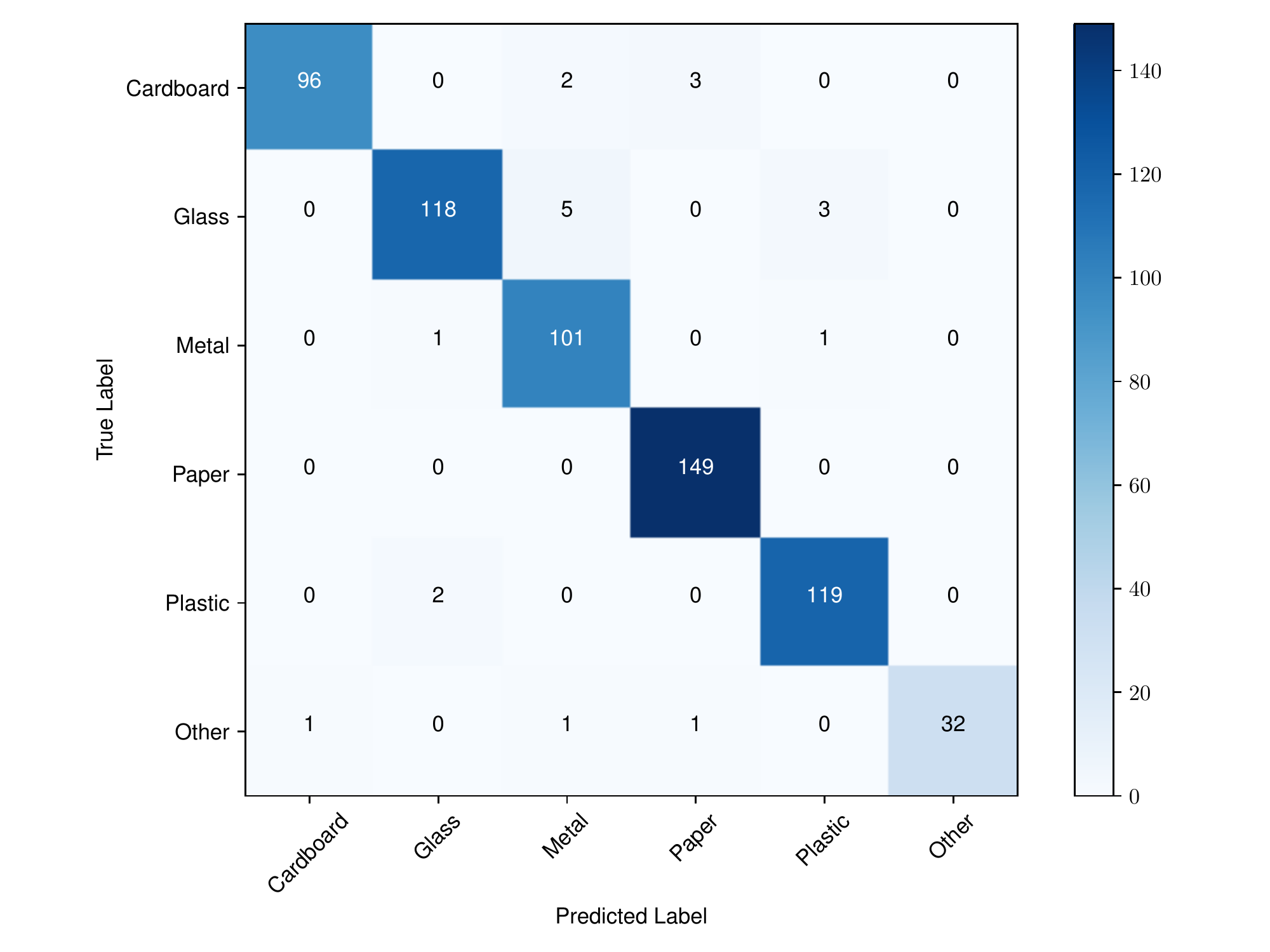}
	\caption{Waste Classification Confusion Matrix}
	\label{fig:confusion_matrix}
\end{figure}

Figure \ref{fig:confusion_matrix} shows the confusion matrix for our WasteNet approach on the test dataset. The y-axis shows the true labels and the x-axis shows the predicted labels. We can see that most of the labels have been correctly predicted as the middle top left to bottom right diagonal has most of the results. Overall, the model has achieved a good prediction accuracy of 0.970. However, we can also see the classes that have been mislabelled in the confusion matrix. The most confused classification is predicting metal when the true label is glass, which happens 5 times. The second most confused classification is also when the material is glass, but the prediction is plastic.

\begin{figure}[!htb]
	\centering
	\subfloat[test][Metal/Glass 9.28/0.00]{\includegraphics[width=.24\textwidth, height=.18\textwidth]{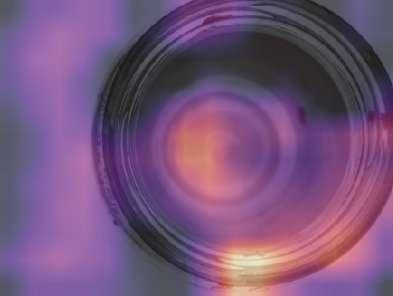}\label{aa}}\hfill
	\subfloat[][Metal/Plastic 18.51/0.00]{\includegraphics[width=.24\textwidth, height=.18\textwidth]{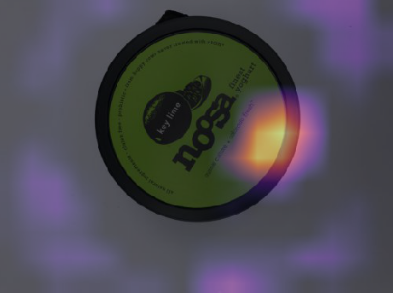}\label{ab}}\hfill\par
	\subfloat[][Paper/Cardboard 6.16/0.00]{\includegraphics[width=.24\textwidth, height=.18\textwidth]{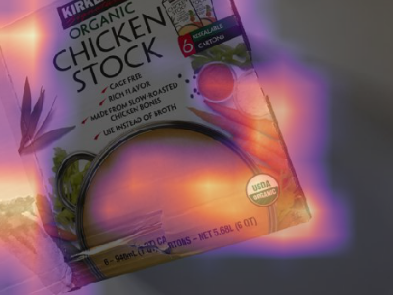}\label{ac}}\hfill
	\subfloat[][Metal/Plastic 3.56/0.03]{\includegraphics[width=.24\textwidth, height=.18\textwidth]{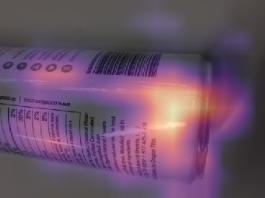}\label{ad}}\par
	\subfloat[][Plastic/Glass 19.17/0.00]{\includegraphics[width=.24\textwidth, height=.18\textwidth]{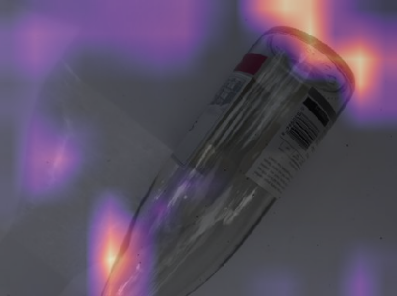}\label{ae}}\hfill
	\subfloat[][Paper/Cardboard 3.92/0.02]{\includegraphics[width=.24\textwidth, height=.18\textwidth]{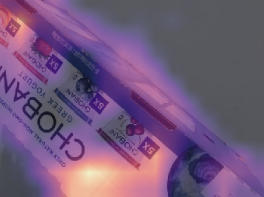}\label{af}}
	\caption{Prediction/Actual Loss/Probability}
	\label{fig:activations}
\end{figure}

To get further insight into how the deep neural network is making the predictions we can use gradient-based localisation to identify the areas in the image that caused the model to make a prediction for a particular class \cite{DBLP:journals/corr/SelvarajuDVCPB16}. Figure \ref{fig:activations} shows a heat map of the regions of the image that cause the network to fire when making the classification decision. These images are taken from the training dataset and are the images that were originally misclassified by the network. Figure \ref{aa} shows an image where the network has made the prediction metal, when the image was actually glass. The network has assigned a probability of 0.00 so was not confident about the prediction. This caused a big change in the network when the true label was revealed, leading to a loss of 9.28. Figure \ref{ae} shows an example where the network has predicted that the bottle is plastic, when it turned out to be glass. We can see that the network has focused on the specific areas to identify the bottle such as the neck and base of the cylinder. As the network is trained on more images in the training dataset it becomes better at classifying the difference between plastic and glass bottles. 

We can see in all these cases that the model had a low probability value close to 0.00. This could be used to avoid these misclassifications in the real-world, where the model would only classify waste when it had a probability of greater than 0.9. Waste with a probability less than this could be separated into a different bin to allow for human sorting on that subset.

\section{Conclusion and Future Work}\label{sect:conclusion}
The results of the experimentation have provided a number of interesting insights for future waste classification. The very high accuracy of 97\% that we are able to achieve shows the potential of these methods. As these models can now be deployed on devices at the edge of the network, such as a Jetson Nano, it allows decision making and artificial intelligence at the edge of the network. The models can be readily deployed in the next versions of smart bins to ensure that the compressed waste is all of the same type. This would increase the amount of waste being recycled, while also reducing the human workload. The method can also be applied to large sorting distribution centres to automatically sort waste into different categories to allow for easier recycling. 

For future work we plan on focusing on waste classification at larger distributions centres, which typically use video feeds. We are planning to updated our model and collect additional data to allow for the automatic classification of waste at a large industrial scale where objects may be overlapping and moving on a conveyor belt. 

\section*{Acknowledgment}
This research has been funded by Science Foundation Ireland (SFI) under grant 13/IA/1885. The Titan V used for this research was donated by the NVIDIA Corporation.

\bibliographystyle{IEEEtran}
\bibliography{ref}
\end{document}